\documentclass[10pt,twocolumn,letterpaper]{article}

\usepackage[pagenumbers]{cvpr}

\usepackage[T1]{fontenc}
\definecolor{cvprblue}{rgb}{0.21,0.49,0.74}
\usepackage[pagebackref,breaklinks,colorlinks,allcolors=cvprblue]{hyperref}

\title{CityRecast: Suppressing Persistent Artifacts in Generative Satellite 3D Reconstruction}

\author{%
  Shuo Cui, Yingyan Hou, Wanxuan Lu\textsuperscript{*}, Hongfeng Yu, Daobing Zhang, Xian Sun\\
  Aerospace Information Research Institute, Chinese Academy of Sciences\\
  {\small \textsuperscript{*}Corresponding author}
}

\begin{document}
\twocolumn[{%
  \renewcommand{\twocolumn}[1][]{#1}% Keep title and teaser in one full-width block.
  \maketitle
  \begin{minipage}{\textwidth}
\centering
\includegraphics[width=\linewidth]{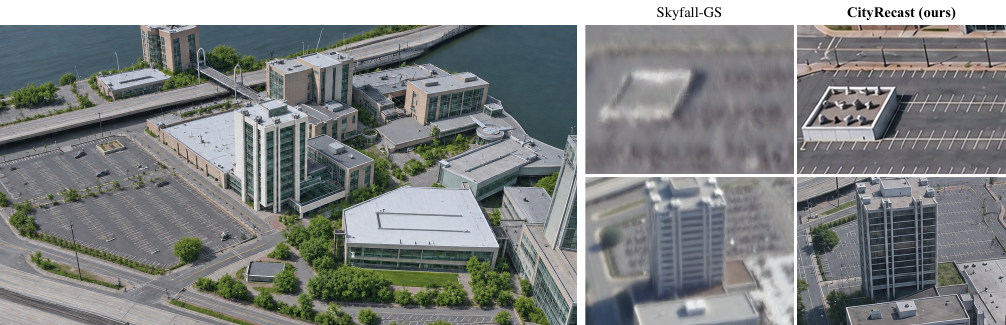}
\captionof{figure}{CityRecast reduces residual artifacts and improves rendering sharpness. Our reconstructed scene is shown on the left, with local comparisons against Skyfall-GS on the right. All images are rendered directly from the final Gaussian scenes.}
\label{fig:teaser}
\end{minipage}

  \vspace{1em}
}]
\begin{abstract}
Satellite-based 3D reconstruction is limited by sparse observations and changes across acquisition dates. Existing methods use generative priors to recover missing visual detail, but ghosting, streaking, and blur persist in their Gaussian scenes. These artifacts become particularly apparent in low-elevation renderings. We introduce CityRecast to suppress these artifacts through Gaussian scene repair and supervision calibration. To correct the Gaussian distributions that produce these artifacts, we first use multi-angle warm-up to constrain their spatial extent. This reduces the initial ghosting carried into progressive refinement. Surface-guided replacement then continues to repair the Gaussian distribution while preserving existing appearance. To provide more reliable supervision for this process, observation calibration mitigates interference from multi-temporal observations. Joint refinement further reduces conflicts among generated views. Experiments on DFC2019 and IARPA demonstrate reduced artifacts in low-elevation renderings while maintaining satellite-view fidelity. Compared with Skyfall-GS, CityRecast reduces FID-CLIP by 25.3\% on DFC2019 and 51.8\% on IARPA. MUSIQ increases by 14.54 and 24.95 points, respectively.
\end{abstract}
    
\section{Introduction}
Reconstructing 3D urban scenes from sparse satellite imagery offers an important route to large-scale city modeling and free-viewpoint visualization~\cite{zhang2019satellite,lit_snerf,lit_satnerf,rs_zhao2023review}. However, limited viewing directions leave building facades weakly observed. Changes in scene content and appearance across acquisition dates further complicate reconstruction~\cite{lit_satnerf,ref09,ref10}. These difficulties become particularly apparent at low-elevation viewpoints unseen during training~\cite{ref18}. Reconstruction should therefore recover clear low-elevation views while preserving fidelity to the available satellite observations.

Multi-view stereo recovers scene geometry by matching satellite observations while accounting for their camera models~\cite{ref01,ref02,ref08}. More recently, neural radiance fields and 3D Gaussian Splatting (3DGS) have enabled joint optimization of scene geometry and appearance through differentiable rendering~\cite{lit_nerf,kerbl2023gaussians}. Their satellite adaptations further account for illumination changes and transient observations~\cite{lit_snerf,lit_satnerf}. However, satellite views still provide limited information about building facades. Generative priors offer a way to recover this missing appearance information~\cite{ref13,ref14}. Skyfall-GS brings this direction to satellite reconstruction by refining rendered views into additional supervision, enabling low-elevation urban exploration~\cite{ref18}.

Despite this progress, persistent artifacts remain in low-elevation renderings. Removing them requires addressing two challenges.
\begin{itemize}
\item Misplaced or excessively elongated Gaussians can reproduce the input satellite views while producing ghosting and streaks from lower elevations. Sharper supervision does not necessarily correct their positions or spatial extent. Repeated fitting smooths some initial artifacts into blur, while others persist as ghosting or streaking.
\item Real and generated images can provide conflicting supervision for the same scene. Transient objects and cross-date appearance changes make real observations inconsistent, while independently refined views may contain incompatible structures. Fitting these targets can reinforce existing artifacts or introduce new ones.
\end{itemize}

To address these challenges, we introduce CityRecast. Our approach repairs the Gaussian scene throughout generative refinement, so that recovered details can be reproduced without retaining the spatial errors behind existing artifacts. This repair also depends on consistent supervision. We therefore reduce conflicts in real and generated observations to guide scene correction and detail recovery.

Our contributions are as follows.
\begin{itemize}
\item We propose CityRecast to suppress persistent artifacts in generative satellite 3D reconstruction. The framework combines Gaussian scene repair with calibration of real observations and joint refinement of generated views, addressing existing scene errors and conflicting supervision together.
\item We develop a stagewise repair strategy that begins with multi-angle warm-up to constrain elongated Gaussians using refined views. Surface-guided replacement transfers their appearance to new surface Gaussians while gradually retiring selected old ones. Observation calibration downweights transient regions and models date-dependent appearance. Geometry-guided joint refinement exchanges cross-view features and aligns structures for consistent supervision.
\item Experiments on six urban scenes demonstrate improved low-elevation rendering quality. CityRecast outperforms Skyfall-GS with the same generator, while surface repair produces clearer renderings than continued fitting with equal GS updates.
\end{itemize}

\begin{figure*}[!t]
\centering
\includegraphics[width=\textwidth]{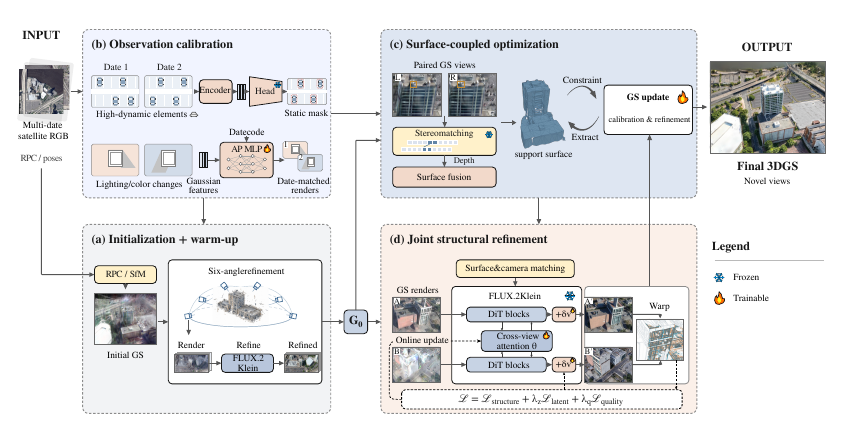}
\caption{Overview of CityRecast. Calibrated satellite images initialize the GS, followed by multi-angle warm-up. During progressive refinement, stereo renderings provide a support surface for Gaussian repair. Jointly refined views supervise GS updates, and the final scene is rendered from novel viewpoints.}
\label{fig:pipeline}
\end{figure*}

\section{Related work}
CityRecast builds on generative priors for satellite reconstruction, with scene initialization and surface-guided repair supported by cross-view refinement. We review the work underlying these components, beginning with satellite reconstruction and generative supervision before turning to geometric guidance and Gaussian optimization.

\noindent\textbf{Satellite scene reconstruction.}
Satellite reconstruction adapts multi-view matching to satellite camera models~\cite{zhang2019satellite,ref08}. Neural radiance fields and 3D Gaussian Splatting offer renderable scene representations~\cite{lit_nerf,kerbl2023gaussians}, with advances in efficiency~\cite{lit_plenoxels,lit_dvgo,lit_tensorf,lit_instantngp} and scale-aware rendering~\cite{lit_mipnerf,lit_mip360,lit_zipnerf,mip2024}. Applying these representations to satellite imagery also requires handling changes in illumination and scene content~\cite{rs_zhang2025brdf}. Shadow Neural Radiance Fields and Satellite Neural Radiance Field model these effects~\cite{lit_snerf,lit_satnerf}, while Season-NeRF addresses seasonal appearance~\cite{ref09}. Recent Gaussian methods further accommodate satellite imaging geometry and sparse, multi-temporal observations~\cite{ref10,ref11,ref12,lit_geogs}.

\noindent\textbf{Generative priors for reconstruction.}
Generative priors supplement incomplete observations by regularizing rendered patches~\cite{lit_diffusionerf} or providing generated and restored targets~\cite{ref13,ref14,lit_gsgs,lit_ri3d}. Correspondence-aware interactions help make these targets consistent across views~\cite{lit_mvdiffusion,lit_epidiff,lit_enhancer3d}. Other approaches encourage consistency through joint view generation~\cite{lit_cat3d,lit_stablecamera} or video priors~\cite{ref15,lit_viewcrafter,ref16,ref17}. In remote sensing, diffusion models also recover occluded object structure through amodal completion~\cite{jrs_lei2026amodal}. In satellite settings, MagicCity uses geometric guidance for city-view generation~\cite{lit_magiccity}, while Skyfall-GS combines observation-based geometry with diffusion priors for progressive low-elevation refinement~\cite{ref18}.

\noindent\textbf{Geometric guidance.}
Classical pipelines combine structure from motion (SfM) with multi-view stereo (MVS) to recover cameras and scene geometry~\cite{ref01,ref02}. Learned MVS improves depth inference~\cite{ref03,lit_patchmatch,lit_iterMVS,lit_transmvs,lit_geomvs,rs_huang2024csc,rs_mao2024sdl}, while DUSt3R, MASt3R, and VGGT extend geometric prediction to pointmaps, correspondences, and cameras~\cite{ref04,ref05,ref06}. Estimated geometry also guides reconstruction through signed distance representations and geometric cues~\cite{lit_neus,lit_volsdf,lit_monosdf,lit_neuralangelo}. Gaussian approaches similarly use geometric constraints for surface recovery~\cite{lit_sugar,lit_twodgs,lit_pgsr,lit_gof,wolf2024gs2mesh}, and geometric estimates for initialization and optimization~\cite{ref07,lit_quicksplat,lit_gsrecon,lit_surfacesplat}.

\noindent\textbf{Sparse and changing observations.}
Sparse-view reconstruction uses learned scene priors to supplement limited observations~\cite{lit_pixelnerf,lit_ibrnet,lit_mvsnerf}. Depth and correspondence supervision provide additional geometric constraints~\cite{lit_dsnerf,lit_sparsenerf,lit_sparf}, while regularization helps reduce overfitting~\cite{lit_dietnerf,lit_regnerf,lit_freenerf}. Gaussian methods improve coverage and geometric constraints~\cite{lit_fsgs,lit_dngaus,lit_scgaussian,lit_comapgs}, or reduce overfitting through pruning and dropout~\cite{lit_sparsegs,lit_dropoutgs}. Multi-temporal imagery captures surface changes under varying observation conditions~\cite{jrs_fu2024timeseries}. These differences complicate fitting a shared scene. NeRF in the Wild models appearance variation and transient content~\cite{lit_nerfw}. Robust fitting and uncertainty weighting instead suppress distractors~\cite{lit_robustnerf,lit_onthego}. Related Gaussian methods also account for appearance changes in uncontrolled captures~\cite{lit_gsw,lit_wildgaussians,lit_vastgaussian}.

\section{Method}
\label{sec:method}

CityRecast suppresses residual artifacts in generative satellite 3D reconstruction by repairing the Gaussian distribution and improving the supervision that guides it.
Fig.~\ref{fig:pipeline} gives an overview. We first define the scene representation (Sec.~\ref{sec:problem_definition}), then describe real and generated supervision (Sec.~\ref{sec:supervision}). Finally, we introduce warm-up and surface-guided Gaussian replacement (Sec.~\ref{sec:gaussian_repair}).

\subsection{Problem definition and notation}
\label{sec:problem_definition}

Given sparse, multi-temporal satellite images and their camera
parameters
$\mathcal{D}=\{(I_t,P_t)\}_{t=1}^{N}$,
our goal is to reconstruct a static Gaussian scene that can be directly
rendered from novel viewpoints.
Following the 3DGS parameterization~\cite{kerbl2023gaussians},
we represent the region of interest as
\begin{equation}
    \begin{aligned}
        \mathcal{G} &= \{g_i\}_{i=1}^{M}, \\
        g_i &= \bigl(
            \boldsymbol{\mu}_i,
            \boldsymbol{\Sigma}_i,
            o_i,
            \mathbf{a}_i,
            \mathbf{h}_i
        \bigr),
    \end{aligned}
    \label{eq:gaussian_representation}
\end{equation}
where
$\boldsymbol{\mu}_i$,
$\boldsymbol{\Sigma}_i$,
$o_i$, and
$\mathbf{a}_i$
denote the center, covariance, opacity, and spherical harmonic (SH)
appearance coefficients, respectively.
We additionally associate each Gaussian with a learnable appearance
feature $\mathbf{h}_i$.

To account for observations acquired on different dates, a shared appearance multilayer perceptron (MLP) conditions Gaussian colors on per-Gaussian features $\mathbf{h}_i$ and per-image appearance codes $\mathbf{e}_t$. Each code captures the appearance conditions of its acquisition date, allowing the same geometry to explain observations with different colors. Real-image fitting uses the corresponding $\mathbf{e}_t$.
Let $\mathcal{R}$ denote the appearance-conditioned differentiable
renderer and $\mathbf{e}_{\mathrm{ref}}$ the reference appearance code.
The renderings at an observed camera $P_t$ and a virtual camera $P_v$
are then
\begin{equation}
    \begin{aligned}
        \widehat{I}_t
        &= \mathcal{R}
        \bigl(\mathcal{G};P_t,\mathbf{e}_t\bigr), \\
        X_v
        &= \mathcal{R}
        \bigl(\mathcal{G};P_v,\mathbf{e}_{\mathrm{ref}}\bigr).
    \end{aligned}
    \label{eq:appearance_conditioned_rendering}
\end{equation}
Real images $I_t$ supervise $\widehat{I}_t$, while generative refinement transforms $X_v$ into additional targets $Y_v$. Both sources supervise the same Gaussian scene $\mathcal{G}$, which must reproduce the recovered details.

The residual artifacts of interest appear in direct renderings of this representation as duplicated road contours and elongated streaks near building boundaries. Floating blobs or sheet-like structures may also appear, along with blurred boundaries and textures.
To address these artifacts, we identify candidate Gaussians for
repair using geometric information and multi-view observations,
and denote their indices by
$\mathcal{B}\subseteq\{1,\ldots,M\}$.
Although these Gaussians may have an inaccurate spatial distribution,
they can still help explain parts of the observed images.
We therefore repair these Gaussians through progressive replacement
rather than immediate removal, using a support surface $\mathcal{S}$
reconstructed from the current scene as a geometric reference.
The repair process is detailed in Sec.~\ref{sec:gaussian_repair}.

\subsection{Calibrating real and generated supervision}
\label{sec:supervision}
\paragraph{Real observation calibration.}
Real satellite images provide primary supervision for Gaussian optimization, but cross-date variations can also disrupt early geometry estimation, including SfM. We therefore address content and appearance changes separately.
Transient objects, such as vehicles, construction equipment, and temporary stockpiles, vary across acquisitions. Fitting them into a shared static scene can leave ghosting or floating artifacts in novel-view renderings.

To mitigate this interference, we adopt transient-aware supervision,
using a pretrained image encoder and prediction head to identify
transient regions.
For pixel $u$ in image $I_t$, the prediction
$p_t^{\mathrm{tr}}(u)\in[0,1]$ denotes the probability that the pixel
belongs to a transient object.
We use this prediction to construct a pixel-wise supervision mask
$m_t(u)$, whose values weight the corresponding photometric errors.
The real-observation fitting loss is
\begin{equation}
    \begin{aligned}
        m_t(u)
        &=
        v_t(u)\bigl[1-p_t^{\mathrm{tr}}(u)\bigr], \\
        \mathcal{L}_{\mathrm{sat}}
        &=
        \frac{
            \sum_{t,u} m_t(u)\,
            \rho\!\left(\widehat{I}_t(u)-I_t(u)\right)
        }{
            \sum_{t,u} m_t(u)
        }.
    \end{aligned}
    \label{eq:transient_aware_supervision}
\end{equation}
Here, $v_t(u)$ denotes the valid-image mask, and $\rho$ measures
photometric error.
Higher transient probabilities reduce the corresponding supervision weights. When the same location is visible in other acquisitions, those observations provide background supervision.

Transient-aware supervision addresses changes in scene content,
but the remaining static regions can still exhibit brightness and
color variations across dates.
We retain the appearance model of Skyfall-GS~\cite{ref18}. A lightweight MLP predicts per-Gaussian color gains and biases. Its inputs are the appearance feature $\mathbf{h}_i$ and the per-image code $\mathbf{e}_t$ encoding acquisition-date conditions. These adjustments are applied before rendering, reducing pressure on shared geometry to explain appearance differences.

\paragraph{Cross-view joint refinement.}
Generatively refined images provide another important source of
supervision for Gaussian reconstruction.
However, structural discrepancies across independently refined
views can introduce conflicting targets for the same scene.
We therefore introduce geometry-guided cross-view feature
aggregation into the frozen FLUX.2 [klein] 4B
backbone~\cite{flux2klein}.
Candidate correspondences are established through
epipolar-constrained feature matching, with support-surface
projections providing additional geometric guidance.
Ray encodings further condition the queries and keys, allowing
attention to incorporate both appearance and spatial relationships.
Bidirectional information exchange preserves the ordering of the
$2\times2$ subcells within each token and updates the generative
features through residual writeback.
For stronger structural control, we add low-rank and linear residuals to each sampling step's native flow prediction. For fused features $\mathbf{f}_{a,k}$ of view $a$ at step $k$, the adapted prediction is
\begin{equation}
    \widehat{\mathbf{v}}_{a,k}
    =
    F_{\mathrm{out}}(\mathbf{f}_{a,k})
    +
    U_1D_1\mathbf{f}_{a,k}
    +
    W_2\mathbf{f}_{a,k},
    \label{eq:joint_flow_adaptation}
\end{equation}
where $F_{\mathrm{out}}$ is frozen. The matrices $D_1$ and $U_1$ form the low-rank residual, while $W_2$ provides the linear residual. Both modify the flow prediction used in the subsequent sampling update.

We jointly refine the two outputs to improve cross-view structural consistency.
To this end, we perform bidirectional image reprojection
between the two target views, then extract multi-scale soft
edges and local contrast.
These cues constrain the alignment of structures such as
window lines and facade openings at valid correspondences.
These cues emphasize structural locations and orientations. An edge-energy lower bound preserves structural detail, while appearance-preservation terms maintain the original colors.

We update the added adaptation parameters online while keeping
the original backbone frozen.
The structural term $\mathcal{L}_{\mathrm{str}}$, latent-space target constraint $\mathcal{L}_z$, and quality-preservation term $\mathcal{L}_q$ define the stage-dependent objective
\begin{equation}
    \mathcal{L}_{\mathrm{joint}}
    =
    \mathcal{L}_{\mathrm{str}}
    +
    \lambda_z\mathcal{L}_z
    +
    \lambda_q\mathcal{L}_q.
    \label{eq:joint_optimization}
\end{equation}
Each update backpropagates through a complete sampling pass. Subsequent view pairs reuse the adapted parameters; the online update schedule is provided in Supplementary Sec.~C.

\begin{figure}[t]
\centering
\includegraphics[width=\linewidth]{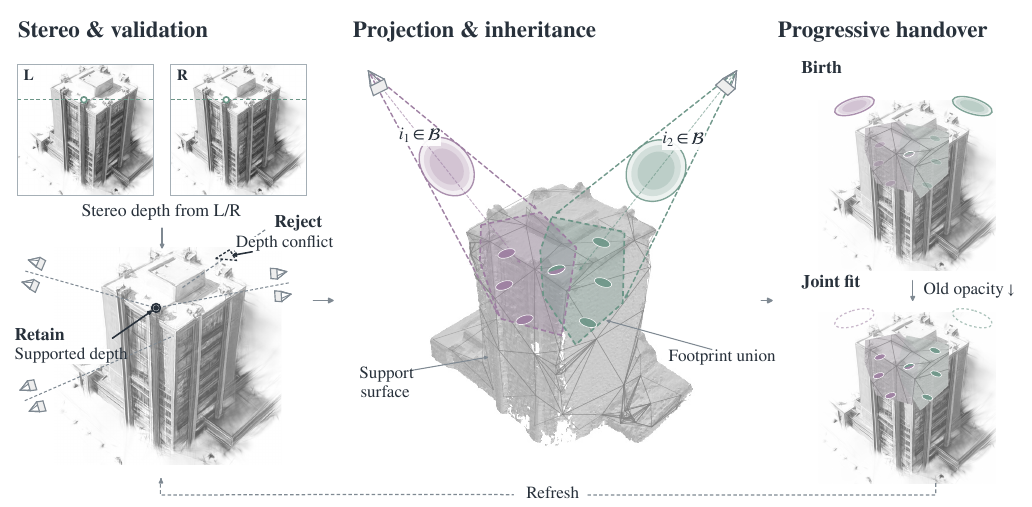}
\caption{Support-surface construction and progressive Gaussian repair. Cross-direction validation forms $\mathcal S$, and projected footprints guide surface-based replacements. Shared colors denote appearance inheritance; fading old Gaussians indicate progressive retirement.}
\label{fig:surface_repair}
\end{figure}

\subsection{Stagewise Gaussian repair}
\label{sec:gaussian_repair}
\paragraph{Multi-angle supervision warm-up.}
We obtain camera parameters and sparse structure through an SfM
pipeline adapted to satellite imaging
geometry~\cite{ref01,zhang2019satellite},
then optimize an initial Gaussian
scene~\cite{kerbl2023gaussians} using the calibrated real satellite
observations.
Under multi-temporal observations captured at high elevations
with limited directional coverage, local image noise and appearance
variations may be absorbed into inaccurate Gaussian configurations
during early fitting.
Spatial extensions along nearly aligned viewing directions remain weakly constrained. These Gaussians can therefore fit the input views with low reconstruction error yet produce pronounced ghosting and streaking in low-elevation renderings.

These artifacts also affect the rendered views used to generate subsequent supervision, allowing errors in the initial scene to be carried into progressive refinement. We therefore intervene before this stage with multi-angle warm-up. Independently refined renderings from multiple azimuths and lower elevations are fitted together with real observations. The additional directions constrain spatial extensions that are weakly observed in the original satellite views. Supplementary Sec.~A provides a local projection analysis. By suppressing these initial artifacts early, warm-up reduces the burden of correcting them during later detail recovery. This prepares a cleaner scene for surface-guided repair, where newly initialized, surface-aligned Gaussians can gradually take over the content previously explained by artifact-producing Gaussians.

% Float scheduling only; method text is unchanged.
% Declared before the experiment section so this wide float can appear on page 6.
\begin{figure*}[!t]
\centering
\includegraphics[width=\linewidth]{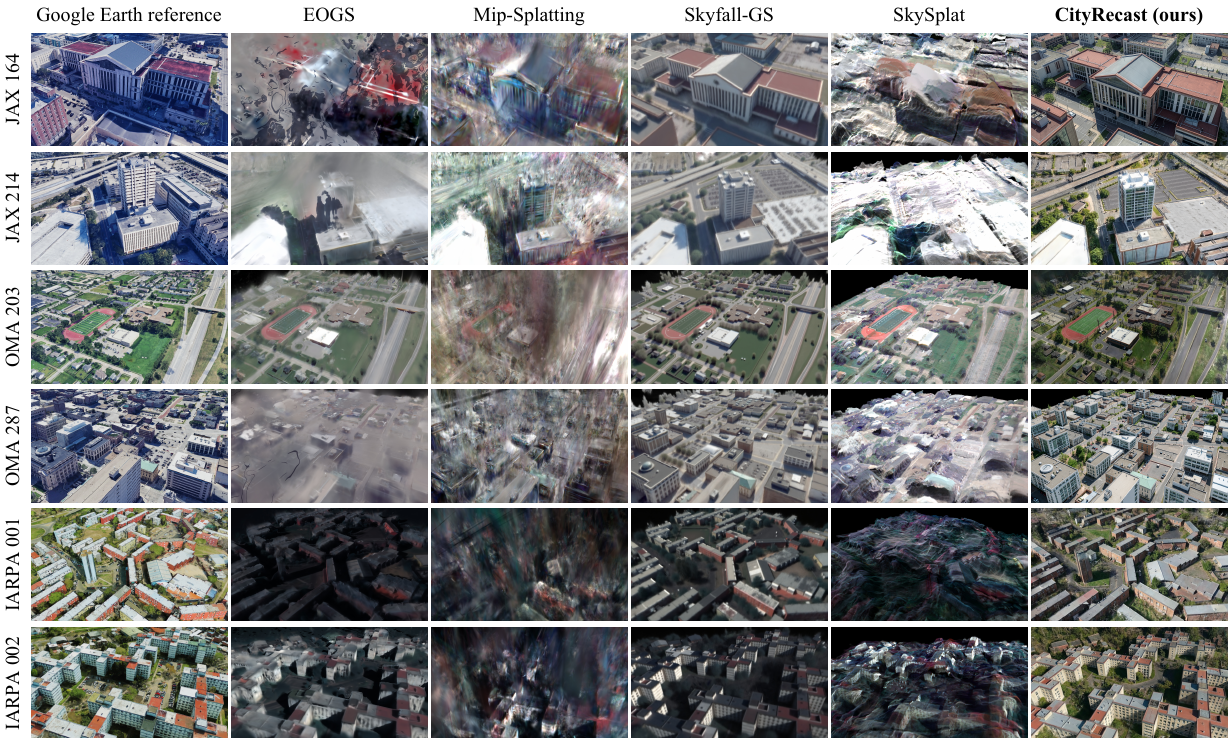}
\caption{Qualitative comparison on six urban scenes. Google Earth images provide visual references. Final GS renderings use matched cameras and crops across methods.}\label{fig:qualitative}
\medskip
\captionof{table}{Quantitative comparison averaged over four DFC2019 and two IARPA scenes. Bold and underline indicate the best and second-best results.}\label{tab:main}
\providecommand{\best}[1]{\textbf{#1}}
\providecommand{\second}[1]{\underline{#1}}
{\small
\setlength{\tabcolsep}{2pt}
\begin{tabular*}{\linewidth}{@{\extracolsep{\fill}}lrrrrrrrrrrrr@{}}
\toprule
& \multicolumn{6}{c}{DFC2019} & \multicolumn{6}{c}{IARPA} \\
\cmidrule(lr){2-7}\cmidrule(l){8-13}
& \multicolumn{3}{c}{Held-out satellite} & \multicolumn{3}{c}{Low-view rendering} & \multicolumn{3}{c}{Held-out satellite} & \multicolumn{3}{c}{Low-view rendering} \\
\cmidrule(lr){2-4}\cmidrule(lr){5-7}\cmidrule(lr){8-10}\cmidrule(l){11-13}
Method & PSNR$\uparrow$ & SSIM$\uparrow$ & LPIPS$\downarrow$ & FID$_{\rm CLIP}\downarrow$ & CMMD$\downarrow$ & MUSIQ$\uparrow$ & PSNR$\uparrow$ & SSIM$\uparrow$ & LPIPS$\downarrow$ & FID$_{\rm CLIP}\downarrow$ & CMMD$\downarrow$ & MUSIQ$\uparrow$ \\
\midrule
EOGS & 16.34 & 0.477 & 0.433 & 54.64 & 3.600 & 48.52 & 19.69 & \second{0.532} & \second{0.276} & 94.01 & 4.899 & 46.19 \\
Mip-Splatting & 15.93 & 0.482 & 0.351 & 73.90 & 4.596 & 41.65 & 18.70 & 0.447 & 0.325 & 102.72 & 6.100 & 29.36 \\
Skyfall-GS & \second{18.87} & \second{0.591} & \second{0.307} & \second{31.22} & \second{2.565} & 52.32 & \second{19.97} & 0.528 & 0.281 & \second{57.90} & \second{3.108} & 43.95 \\
SkySplat & 10.91 & 0.315 & 0.509 & 65.71 & 4.646 & \second{53.19} & 15.68 & 0.263 & 0.483 & 87.25 & 4.863 & \second{52.55} \\
\midrule
\textbf{CityRecast} & \best{19.21} & \best{0.613} & \best{0.295} & \best{23.33} & \best{1.860} & \best{66.86} & \best{20.23} & \best{0.552} & \best{0.264} & \best{27.91} & \best{1.885} & \best{68.90} \\
\bottomrule
\end{tabular*}}

\end{figure*}

% Keep the generator control near the transition to the experiments.
\begin{figure}[!t]
\centering
\includegraphics[width=\linewidth]{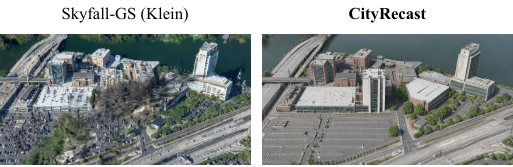}
\caption{Comparison with the same refinement model. Both methods use Klein~4B. CityRecast recovers facade detail that remains obscured by streaks in Skyfall-GS.}\label{fig:generator_control}
\end{figure}

\paragraph{Support surface construction and validation.}
Fig.~\ref{fig:surface_repair} illustrates how support-surface construction leads to Gaussian replacement.
Multi-angle warm-up helps reduce initial ghosting, but local
inconsistencies may persist in subsequent real and generated
supervision.
Supervision-driven densification also does not explicitly constrain
new Gaussians to plausible scene surfaces.
We therefore update a support surface throughout reconstruction
to guide Gaussian initialization and spatial repair.

We estimate depth from calibrated stereo pairs rendered from the current Gaussian scene~\cite{wolf2024gs2mesh}. Unlike per-view monocular prediction, stereo uses known camera baselines to recover geometry at the scene's scale and supports correspondence consistency checks. We render pairs at multiple azimuths and elevations to improve coverage under occlusion.

However, these stereo inputs still originate from the current
Gaussian scene.
Rendering artifacts, particularly during early reconstruction,
may therefore be reconstructed as spurious surfaces.
We validate candidate surface points against depth estimates from other directions. Using a validity mask $w_v(\mathbf{x})$ that combines left--right consistency, $\alpha$ coverage in both images, and projection validity, we aggregate geometric support and free-space conflicts as
\begin{equation}
    \begin{aligned}
        s(\mathbf{x})
        &=
        \sum_v w_v(\mathbf{x})\,
        \mathbf{1}\!\left[
            |\Delta_v(\mathbf{x})| \leq \tau_v
        \right], \\
        c(\mathbf{x})
        &=
        \sum_v w_v(\mathbf{x})\,
        \mathbf{1}\!\left[
            \Delta_v(\mathbf{x}) > \tau_v
        \right].
    \end{aligned}
    \label{eq:surface_support_conflict}
\end{equation}
Here, $\Delta_v(\mathbf{x})=D_v(\pi_v(\mathbf{x}))-z_v(\mathbf{x})$ compares estimated surface depth $D_v$ with candidate depth $z_v$ at projection $\pi_v(\mathbf{x})$. With depth tolerance $\tau_v$, $\Delta_v>\tau_v$ places the candidate in front of the surface, indicating free-space conflict. Occluded or invalid observations contribute no conflicts.

\paragraph{Projection-based surface Gaussian initialization and progressive handover.}
Selected Gaussians may still explain valid observations, so direct removal risks losing useful content, while abrupt relocation changes coverage and occlusion. We instead initialize surface Gaussians from their projected footprints and transfer their multi-view appearance before progressive handover.

For each candidate $i\in\mathcal{B}$, we intersect rays sampled within its projected ellipsoidal footprint $\pi_t(E_i)$ with $\mathcal{S}$. We then merge valid, visible intersections across candidates and views, deduplicate them, and sample new Gaussian centers under density control:
\begin{equation}
\begin{aligned}
\mathcal{F}_{it}
&=\operatorname{Hit}_{\mathcal{S}}\!\left(P_t,\pi_t(E_i)\right),\\
\{\boldsymbol{\mu}_j^0\}_j
&=\operatorname{Sample}_{\mathrm{dens}}\!\left(\bigcup_{i\in\mathcal{B}}\bigcup_t\mathcal{F}_{it}\right).
\end{aligned}
\label{eq:surface_footprint}
\end{equation}
This preserves the candidate's full projected coverage rather than using its center alone.

At each sampled position, we initialize a surface-aligned Gaussian with small normal thickness and bounded tangential scales. The contributing Gaussian--view pairs $\mathcal{P}_j$ provide directional colors, which we use to fit the initial SH appearance
\begin{equation}
    \mathbf{a}_j^{0}
    \;\leftarrow\;
    \underset{\mathbf{a}}{\arg\min}
    \sum_{(i,t)\in\mathcal{P}_j}
    \omega_{ijt}
    \left\|
        \operatorname{SH}(\mathbf{a},\mathbf{d}_{jt})
        -
        c_i(\mathbf{d}_{it})
    \right\|_2^2.
    \label{eq:appearance_inheritance}
\end{equation}
Here, $\mathbf{d}_{it}$ and $\mathbf{d}_{jt}$ denote source and new Gaussian viewing directions in camera $t$. The fit uses source color $c_i(\mathbf{d}_{it})$ and coverage weight $\omega_{ijt}$.

We optimize new and existing Gaussians together while gradually lowering the opacity cap of selected old Gaussians, allowing surface Gaussians to take over. The shared rendering objective is detailed in Supplementary Sec.~B. At designated repair stages, we reconstruct and validate the support surface and refresh $\mathcal{B}$ from the updated scene. Improved renderings thus provide better geometric guidance for subsequent repair.

\providecommand{\oursblank}{}
\providecommand{\baselineblank}{--}
\providecommand{\best}[1]{\textbf{#1}}
\providecommand{\second}[1]{\underline{#1}}

\section{Experiments}\label{sec:experiments}
We first evaluate CityRecast on DFC2019 and IARPA against existing reconstruction methods, assessing low-view rendering quality and satellite-view fidelity (Sec.~\ref{sec:comparisons}). We further compare with Skyfall-GS using the same refinement model to examine the contribution of our reconstruction strategy. Finally, we ablate individual components to analyze their effects on rendering quality and residual artifacts (Sec.~\ref{sec:ablation}).

\subsection{Experimental setup}\label{sec:experimental_setup}
\paragraph{Evaluation protocol.}
We reconstruct multiple urban scenes from DFC2019 and IARPA. Each scene is evaluated from 18 low-view cameras spanning $25^\circ$, $35^\circ$, and $45^\circ$ at six azimuths. All methods use the same cameras, resolution, and crops, and are evaluated by directly rendering their final Gaussian scenes. Our comparisons include EOGS~\cite{ref10}, Mip-Splatting~\cite{mip2024}, Skyfall-GS~\cite{ref18}, and SkySplat~\cite{ref11}. 

PSNR, SSIM, and LPIPS measure fidelity to held-out satellite observations. For low-view renderings, we report MUSIQ~\cite{ke2021musiq} and compute FID-CLIP and CMMD using Google Earth imagery as distributional references. Per-scene scores are averaged equally within each dataset.

\paragraph{Implementation details.}
We first fit the initial Gaussians to real observations, then use multi-angle warm-up to prepare the scene for progressive refinement. Progressive refinement follows the curriculum principle of Skyfall-GS~\cite{ref18}, gradually lowering the viewing elevation. A final consolidation stage, like warm-up, mixes refined targets from multiple elevations to consolidate details across viewing directions. Real observations remain part of the supervision throughout training. Detailed schedules and sampling proportions are provided in Supplementary Sec.~C. Each scene takes approximately 5--8 hours to train on one PRO~6000 GPU.

% Declare upcoming floats early to spread them across the experiment pages.
\begin{figure}[!t]
\centering
\includegraphics[width=\linewidth]{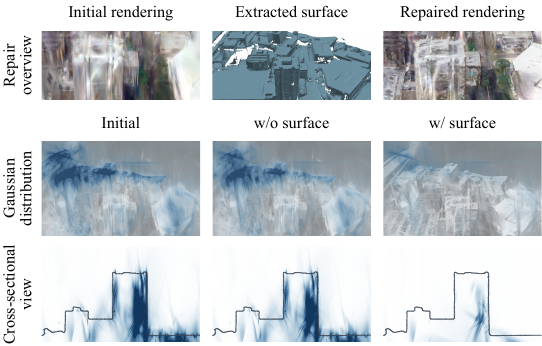}
\caption{Effect of surface-guided repair. The top row shows the initial rendering, extracted surface, and repaired rendering. The middle row compares the spatial extent of artifacts in the initial scene, after continued fitting, and after surface repair. The bottom row shows the corresponding cross sections within a 40\,m slab. Candidates and display scales are shared across columns.}\label{fig:surface_ablation}\label{fig:surface_rgb}
\end{figure}

\begin{figure}[!t]
\centering
\includegraphics[width=\linewidth]{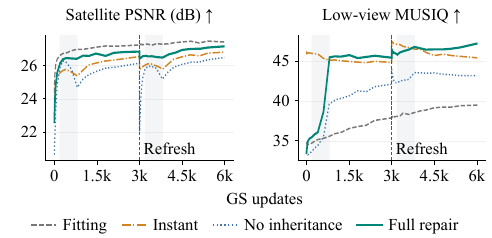}
\caption{Repair dynamics over 6k GS updates. Fitting denotes continued optimization. Instant uses immediate handover, whereas No inheritance omits appearance transfer. Dashed vertical lines indicate when the surface is refreshed.}\label{fig:repair_dynamics}
\end{figure}

% Author-supplied ablation scores; preserve numerical entries.
% Rows (a)-(d): without W, C, S, and J, respectively.
% C jointly enables transient weighting and date-conditioned appearance.
% Without J, use independent Klein refinement with the same backbone.
\begin{table}[!t]
\centering
\caption{Ablation study on JAX164 and OMA203. Scores are averaged over the two scenes. W, C, S, and J denote warm-up, observation calibration, surface repair, and joint refinement, respectively.}\label{tab:ablations}
{\footnotesize
\renewcommand{\arraystretch}{1.12}
\setlength{\tabcolsep}{1pt}
\begin{tabular*}{\columnwidth}{@{\extracolsep{\fill}}lcccc rrrr@{}}
\toprule
& \multicolumn{4}{c}{Components} & \multicolumn{2}{c}{Satellite} & \multicolumn{2}{c}{Low-view} \\
\cmidrule(lr){2-5}\cmidrule(lr){6-7}\cmidrule(l){8-9}
Method & W & C & S & J & PSNR$\uparrow$ & LPIPS$\downarrow$ & FID$_{\rm CLIP}\downarrow$ & MUSIQ$\uparrow$ \\
\midrule
(a) & -- & $\checkmark$ & $\checkmark$ & $\checkmark$ & 18.13 & 0.379 & 30.73 & 57.33 \\
(b) & $\checkmark$ & -- & $\checkmark$ & $\checkmark$ & 17.29 & 0.411 & 27.87 & 61.02 \\
(c) & $\checkmark$ & $\checkmark$ & -- & $\checkmark$ & 18.42 & 0.327 & 28.93 & \second{63.58} \\
(d) & $\checkmark$ & $\checkmark$ & $\checkmark$ & -- & \best{18.94} & \best{0.296} & \second{27.46} & 59.19 \\
\midrule
\textbf{Full} & $\checkmark$ & $\checkmark$ & $\checkmark$ & $\checkmark$ & \second{18.57} & \second{0.317} & \best{23.36} & \best{67.42} \\
\bottomrule
\end{tabular*}}
\end{table}

\subsection{Comparison with existing methods}\label{sec:comparisons}
\paragraph{Quantitative and qualitative results.}
Table~\ref{tab:main} reports results on DFC2019 and IARPA. CityRecast outperforms the comparison methods on all six metrics in both datasets. Compared with Skyfall-GS, it reduces FID-CLIP by 25.3\% on DFC2019 and 51.8\% on IARPA, with lower CMMD and higher MUSIQ on both. Satellite-view metrics show modest improvements, indicating that CityRecast maintains satellite-view reconstruction fidelity while improving low-view rendering quality.

Fig.~\ref{fig:qualitative} compares direct Gaussian renderings across six scenes. CityRecast substantially reduces visible artifacts and blur relative to all four baselines, producing sharper building boundaries and clearer facade details. EOGS and Mip-Splatting exhibit smeared surfaces and elongated artifacts that obscure scene content, while SkySplat shows rippled surfaces and fragmented structures. Skyfall-GS recovers more coherent scene layouts, but facade patterns and roof edges remain visibly blurred. In JAX~214, CityRecast reveals more distinct vertical facade divisions and sharper roof edges around the tower. Roof outlines and gaps between buildings are also clearer in OMA~287, while the IARPA scenes show better separation of buildings, vegetation, and surrounding ground. The Google Earth references also help identify structures obscured in the baselines, such as the sports field and nearby buildings in OMA~203. CityRecast renders their boundaries more distinctly. 

\paragraph{Comparison with the same generator.}
To examine the contribution of the reconstruction strategy, we replace the original Skyfall-GS refiner with the FLUX.2 [klein]~4B used by CityRecast. The rest of the Skyfall-GS pipeline remains unchanged. Fig.~\ref{fig:generator_control} shows that residual streaks still obscure parts of the facade in this control. CityRecast reveals clearer structural detail with the same generator, illustrating the benefit of its reconstruction strategy.

\subsection{Ablation and analysis}\label{sec:ablation}
Table~\ref{tab:ablations} evaluates the contribution of each component by removing it from the full pipeline. Observation calibration is ablated as a whole, including both transient weighting and date-conditioned appearance modeling.

% Author-confirmed control: replace the 10k warm-up with 10k progressive-refinement updates at 85 degrees.
% This matches GS update count, not total wall-clock or generator compute.
% Disabling J removes joint refinement and its online adaptation together.

\paragraph{Multi-angle warm-up.}
For row (a), we replace multi-angle warm-up with high-elevation refinement using the same number of GS updates. Both variants then follow the same subsequent training schedule. The replacement configuration is provided in Supplementary Sec.~C. Without warm-up, FID-CLIP increases from 23.36 to 30.73 and MUSIQ drops from 67.42 to 57.33, the largest degradation in low-view quality among the ablations. PSNR and LPIPS also deteriorate. With the same number of GS updates, multi-angle warm-up provides a better starting point for subsequent training than continued fitting at high elevation. Supplementary Sec.~D provides a separate comparison with continued fitting under matched GS-update budgets, showing RGB renderings and digital surface model (DSM) visualizations.

\paragraph{Observation calibration.}
Removing calibration lowers PSNR from 18.57 to 17.29~dB and raises LPIPS from 0.317 to 0.411, with a corresponding deterioration in low-view quality. A related failure is visible in the parking areas of Fig.~\ref{fig:generator_control}, where the Skyfall-GS control contains dense ghosting and streaks extending into nearby structures. Vehicles present on different dates can be absorbed into the GS as persistent content. Subsequent generative refinement can reinforce and reshape these remnants, spreading artifacts beyond their original locations. CityRecast instead shows cleaner parking surfaces, distinct lane markings, and less clutter around the buildings. Additional examples of transient observation calibration are shown in Supplementary Sec.~D.

\paragraph{Surface-guided repair.}
Removing surface repair raises FID-CLIP from 23.36 to 28.93, while PSNR decreases by only 0.15~dB (Table~\ref{tab:ablations}). This difference is more evident in Fig.~\ref{fig:surface_ablation}. Although the satellite-view fitting scores change only modestly, surface repair substantially improves the low-view appearance. Starting from the same initial GS, we compare continued fitting with surface repair after 1,800 additional GS updates in each case. Continued fitting leaves elongated Gaussian structures and the resulting veiling largely intact, whereas surface repair reduces their spatial extent and reveals clearer facade detail. At the same update budget, surface repair reduces residual artifacts more effectively than continued fitting.

Fig.~\ref{fig:repair_dynamics} further examines appearance inheritance and progressive handover. Without appearance inheritance, surface refresh causes a pronounced drop in satellite-view PSNR. Full repair limits this disruption and achieves the highest low-view MUSIQ by the end of optimization. Continued fitting retains slightly higher PSNR but yields substantially lower low-view quality.

\begin{figure}[!tb]
\centering
\includegraphics[width=\linewidth]{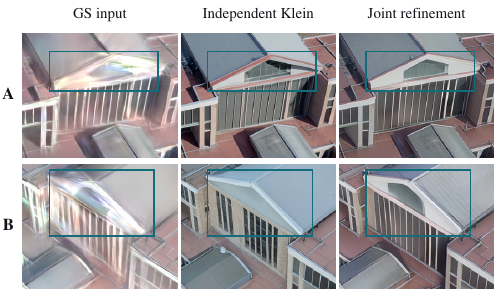}
\caption{Effect of joint refinement on generated supervision. Views A and B compare independent Klein outputs with jointly refined targets after ten online updates; highlighted regions mark corresponding roof structures.}\label{fig:joint}
\end{figure}

\paragraph{Cross-view refinement.}
 To ablate J, we use independent Klein~4B outputs without online adaptation, keeping the view count and sampling steps unchanged. Joint refinement improves low-view FID-CLIP and MUSIQ, accompanied by a slight decrease in satellite-view fidelity. Given our focus on low-elevation rendering, we favor this improvement in perceptual quality. Fig.~\ref{fig:joint} illustrates the effect on generated supervision. Independent refinement produces incompatible roof structures in the two views, whereas joint refinement better aligns the highlighted structures after ten online updates.

\section{Conclusion}\label{sec:conclusion}
CityRecast suppresses persistent artifacts in generative satellite reconstruction through early multi-angle correction and continued surface-guided repair. Observation calibration and joint refinement support this repair by reducing conflicts in real and generated supervision. Experiments across six urban scenes show reduced ghosting, streaking, and blur in direct GS renderings. Building boundaries and facade details become clearer, while satellite-view fidelity is maintained. These results highlight the importance of repairing the underlying Gaussian scene alongside improving generated supervision. However, visually plausible facades in weakly observed regions may still differ from the actual buildings. Future work will focus on reducing runtime and improving the faithfulness of low-elevation facade reconstruction to real structures.

{
    \small
    \bibliographystyle{ieeenat_fullname}
    \bibliography{main}
}

\end{document}